\documentclass[11pt,a4paper]{article}
\usepackage[hyperref]{acl2017}
\usepackage{times}
\usepackage{latexsym}
\usepackage{graphicx}
\usepackage{url}
\usepackage{amsmath}
\usepackage{adjustbox}
\usepackage[ngerman,english]{babel}
\usepackage{tabularx}
\usepackage{booktabs}
\usepackage{todonotes}
\usepackage{placeins}

\aclfinalcopy % Uncomment this line for the final submission
\title{On Extending Neural Networks with Loss Ensembles for Text Classification}

\author{Hamideh Hajiabadi \\
  Ferdowsi University of Mashhad (FUM)\\
  Mashhad, Iran \\
  {\tt \small hamideh.hajiabadi@mail.um.ac.ir} \\\And
  Diego Molla-Aliod \\
  Macquarie University\\
  Sydney, New South Wales, Australia \\  
  {\tt \small diego.molla-aliod@mq.edu.au} \\\AND
  Reza Monsefi \\
  Ferdowsi University of Mashhad (FUM)\\
  Mashhad, Iran\\  
  {\tt monsefi@um.ac.ir} \\}

\begin{document}
\maketitle
\begin{abstract}
  Ensemble techniques are powerful approaches that combine several weak learners to build a stronger one. As a meta learning framework, ensemble techniques can easily be applied to many machine learning techniques. In this paper we propose a neural network extended with an ensemble loss function for text classification. The weight of each weak loss function is tuned within the training phase through the gradient propagation optimization method of the neural network. The approach is evaluated on several text classification datasets. We also evaluate its performance in various environments with several degrees of label noise. Experimental results indicate an improvement of the results and strong resilience against label noise in comparison with other methods.  
\end{abstract}

\section{Introduction}

In statistics and machine learning, ensemble methods use multiple learning algorithms to obtain better predictive performance \cite{mannor2001weak}. It has been proved that ensemble methods can boost weak learners whose accuracies are slightly better than random guessing into arbitrarily accurate strong learners \cite{bai2014bayesian,zhang2016bayesian}. When it could not be possible to directly design a strong complicated learning system, ensemble methods would be a possible solution. In this paper, we are inspired by ensemble techniques to combine several weak loss functions in order to design a stronger ensemble loss function for text classification.

In this paper we will focus on multi-class classification where the class to predict is encoded as a vector $y$ with the one-hot encoding of the target label, and the output of a classifier $\hat{y} = f(x;\theta)$ is a vector of probability estimates of each label given input sample $x$ and training parameters $\theta$. Then, a loss function $L(y ,\hat{y})$ is a positive function that measures the error of estimation \cite{steinwart2008support}. Different loss functions have different properties, and some well-known loss functions are shown in Table~\ref{loss functions}. %For example, minimization of the Square Loss leads to the Mean Square Estimator (MSE) that performs well with Gaussian noise, while minimizing the Zero-One Loss leads to the Maximum A Posteriori (MAP) estimator which has quite good performance in applications for which margin is not important \cite{uhlich2012bayesian,wang2010ssim}. 
Different loss functions lead to different Optimum Bayes Estimators having their own unique characteristics. %For example, while Absolute Loss works well on data with Laplace noise, Square Loss is suitable for data with Gaussian noise. 
So, in each environment, picking a specific loss function will affect performance significantly \cite{xiao2017ramp,zhao2010convex}.    

\begin{table*}[ht]
\begin{tabular}{p{0.5\linewidth} |p{0.5\linewidth}}%{*{p}{l}}
\hline
 \bf Name of loss function & \bf  $L(y,\hat{y})$  \\
 \hline
Zero-One
 \cite{xiao2017ramp}&$L_{0-1} = 
      \begin{cases} 
      0 & z\geq 0 \\
      1 & z < 0     
   \end{cases}$ \\
Hinge Loss
\cite{masnadi2009design,steinwart2002support} & $L_{H} = 
      \begin{cases} 
      0 & z\geq 1 \\
      \max(0,1-z) & z < 1     
   \end{cases}$ \\
   
Smoothed Hinge 
\cite{zhao2010convex} & $L_{S-H}=\begin{cases}
0&z\geq 1\\
\frac{1-z^2}{2}& 0 \leq z < 1\\
\max(0,1-z)& z\leq 0
\end{cases} $ \\

 Square Loss 
 & $L_{S}=\|y-\hat{y}\|_2^2$  \\
 Correntropy Loss \cite{4355325,1716783}
 & $L_{C}=\exp\frac{\|y-\hat{y}\|_2^2}{\sigma^2}$  \\
Cross-Entropy Loss 
\cite{masnadi2010design} & $L_{C-E}=\log{(1+\exp{(-z)})}$  \\
Absolute Loss 
& $L_{A}=\|y-\hat{y}\|_1 $\\
\hline
\end{tabular}

\caption{\label{loss functions}Several well-known loss functions, where $z=y\cdot\hat{y} \in {\cal R}$. }
\end{table*}

In this paper, we propose an approach for combining loss functions which performs substantially better especially when facing annotation noise. The framework is designed as an extension to regular neural networks, where the loss function is replaced with an ensemble of loss functions, and the ensemble weights are learned as part of the gradient propagation process. We implement and evaluate our proposed algorithm on several text classification datasets.

The paper is structured as follows. An overview of several loss functions for classification is briefly introduced in Section~\ref{sec:background}. The proposed framework and the proposed algorithm are explained in Section~\ref{sec:proposed}. Section~\ref{sec:experiments} contains experimental results on classifying several text datasets. The paper is concluded in Section~\ref{sec:conclusions}.   

\section{Background}\label{sec:background}

A typical machine learning problem can be reduced to an expected loss function minimization problem \cite{bartlett2006convexity,painsky2016isotonic}. \newcite{rosasco2004loss} studied the impact of choosing different loss functions from the viewpoint of statistical learning theory. In this section, several well-known loss functions are briefly introduced, followed by a review of ensemble methods. 

In the literature, loss functions are divided into margin-based and distance-based categories. Margin-based loss functions are often used for classification purposes \cite{steinwart2008support,khan2013semi,chen2017kernel}. Since we evaluate our work on classification of text datasets, in this paper we focus on margin-based loss functions. 

%\subsection{Margin-Based Loss Functions}
A margin-based loss function is defined as a penalty function $L(y, \hat{y})$  based in a margin $z = y\cdot \hat{y}$. 
In any given application, some margin-based loss functions might have several disadvantages and advantages and we could not certainly tell which loss function is preferable in general. For example, consider the Zero-One loss function which penalizes all the misclassified samples with the constant value of 1 and the correctly classified samples with no loss. This loss function would result in a robust classifier when facing outliers but it would have a terrible performance in an application with margin focus \cite{zhao2010convex}. 

A loss function is margin enforcing if minimization of the expected loss function leads to a classifier enhancing the margin \cite{masnadi2009design}. Learning a classifier with an acceptable margin would increase generalization. Enhancing the margin would be possible if the loss function returns a small amount of loss for the correct samples close to the classification hyperplane. For example, Zero-One does not penalize correct samples at all and therefore it does not enhance the margin, while Hinge Loss is a margin enhancing loss function.

The general idea of ensemble techniques is to combine different expert ideas aiming at boosting the accuracy based on enhanced decision making. Predominantly, the underlying idea is that the decision made by a committee of experts is more reliable than the decision of one expert alone \cite{bai2014bayesian,mannor2001weak}. Ensemble techniques as a framework have been applied to a variety of real problems and better results have been achieved in comparison to using a single expert.   

Having considered the importance of the loss function in learning algorithms, in order to reach a better learning system, we are inspired by ensemble techniques to design an ensemble loss function. The weight applied to each weak loss function is tuned through the gradient propagation optimization of a neural network working on a text classification dataset. 

Other works \cite{Shi2015,BenTaieb2016} have combined two loss functions where the weights are specified as a hyperparameter set prior to the learning process (e.g. during a fine-tuning process with crossvalidation). In this paper, we combine more than two functions and the hyperparameter is not set a-priory but it is learned during the training process.

%In the next section, the proposed approach is fully elaborated. 

\section{Proposed Approach}\label{sec:proposed}
Let $(x,y)$ be a sample where $x \in {\cal R}^N$ is the input and $y \in \{0,1\}^C$ is the one-hot encoding of the label ($C$ is the number of classes). Let $\theta$ be the parameters of a neural network classifier with a top softmax layer so that the probability estimates are $\hat{y}=softmax(f(x;\theta))$. Let $\{L_i(y,\hat{y})\}_{i=1}^M$ denote $M$ weak loss functions. In addition to finding the optimal $\theta$, the goal is to find the best weights , $\{\lambda_1,\lambda_2,\dots,\lambda_M\}$, to combine $M$ weak loss functions in order to generate a better application-tailored loss function. We need to add a further constraint to avoid yielding near zero values for all $\lambda_i$ weights. The proposed ensemble loss function is defined as below.
\begin{equation}
L=\sum_{j=1}^M \lambda_j L_j(y,\hat{y}),\sum_{j=1}^M \lambda_j =1
\end{equation}
The optimization problem could be defined as follows, given $T$ training samples.
\begin{equation}
\begin{aligned}
 \underset{\theta,\lambda}{\text{minimize}}
 & \sum_{i=1}^T\sum_{j=1}^M \lambda_j L_j(y_i,\hat{y}_i) \\
  \text{s.t.}
 & \sum_{j=1}^M \lambda _j=1,\lambda_i\geq 0
\end{aligned}
\end{equation}
To make the optimization algorithm simpler, we use $\lambda_i^2 $ instead of $\lambda_i$, so the second constraint $\lambda _i \geq 0$ can be omitted.  We then incorporate the constraint as a regularization term based on the concept of Augmented Lagrangian. The modified objective function using Augmented Lagrangian is presented as follows.
\begin{equation}\label{formula}
\begin{aligned}
 \underset{\theta,\lambda}{\text{minimize}}
 &\sum_{i=1}^T\sum_{j=1}^M \lambda_j^2 L_j(y_i,\hat{y}_i)+ &\\
 & \eta_1 (\sum_{j=1}^M \lambda _j^2-1)+\eta_2 (\sum_{j=1}^M \lambda _j^2-1)^2\\
\end{aligned}
\end{equation}
Note that the amount of $\eta_2$ must be significantly greater that $\eta_1$ \cite{nocedal2006penalty} . The first and the second terms of the objective function cause $\lambda_i^2$ values to approach zero but the third term satisfies $\sum_{j=1}^M \lambda _j^2=1$. 

\begin{figure}
\centering
\includegraphics[ width=0.5\textwidth]{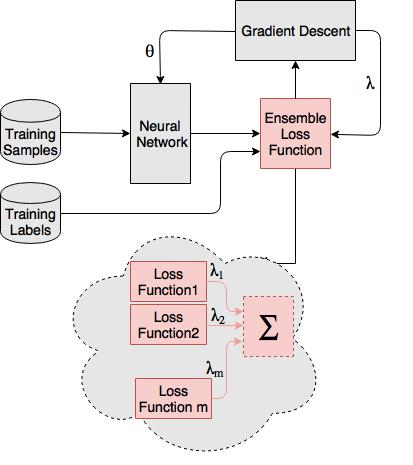}
\caption{ \label{diagram} The proposed learning diagram}
\end{figure}

Figure~\ref{diagram} illustrates the framework of the proposed approach with the dashed box representing the contribution of this paper.  In the training phase, the weight of each weak loss function is trained through the gradient propagation optimization method. The accuracy of the model is calculated in a test phase not shown in the figure. 

\section{Experimental Results}\label{sec:experiments}
We have applied the proposed ensemble loss function to several text datasets. Table \ref{datasets} provides a brief description of the datasets. To reach a better ensemble loss function we choose three loss functions with different approaches in facing with outliers, as weak loss functions: Correntropy Loss which does not assign a high weight to samples with big errors, Hinge Loss which penalizes linearly and Cross-entropy Loss function which highly penalizes the samples whose predictions are far from the targets. We compared results with 3 loss functions which are widely used in neural networks: Cross-entropy, Square Loss, and Hinge Loss. %The following two constraints are added to prevent $\lambda_i$ approaching zero:

%\begin{equation}
%\begin{aligned}
% \underset{\theta,\lambda}{\text{min}}
% &\sum_{i=1}^T \lambda_1^2 L_1+\lambda_2^2 L_2+\lambda_3^2 L_3+ &\\
% & \eta_1 (\lambda _1^2+\lambda _2^2+\lambda _3^2-1)+\eta_2 (\lambda _1^2+\lambda _2^2+\lambda _3^2-1)^2\\
%\end{aligned}
%\end{equation}
% The first and second terms of the objective function cause $\lambda_i$ approaching zero and the third one satisfies $\sum_{j=1}^M \lambda _j^2=1$. 
We picked $\eta_1$ near zero and $\eta_2=200$ in~(\ref{formula}). 
 
\begin{table}[ht]
\centering
\small
\begin{tabularx}{\columnwidth}{X X} 
\hline \bf Name of Datasets& \bf Description \\ \hline
20-newsgroup &
This data set is a collection of
20,000 messages,collected from
20 different net-news newsgroups. \\ 
Movie-reviews in corpus & The NLTK corpus movie-reviews data set has the reviews, and they are labeled already as positive or negative. 
\\
Email-Classification (TREC)& It is a collection of sample emails (i.e. a text corpus). In this corpus, each email has already been labeled as Spam or Ham. \\
Reuters-21578& 
The data was originally collected
and labeled by Carnegie Group, 
Inc. and Reuters, Ltd. in the 
course of developing the 
CONSTRUE text categorization 
system \\
\hline
\end{tabularx}
\caption{\label{datasets} Description of dataset }
\end{table}
%We compare the proposed approach with several other methods on the datasets in Table~\ref{datasets}. 
Since this work is a proof of concept, the neural networks of each application are simply a softmax of the linear combination of input features plus bias:
$$
\hat{y} = \hbox{softmax}(x\cdot W + b)
$$
where the input features $x$ are the word frequencies in the input text. Thus, $\theta$ in our notation is composed of $W$ and $b$. We use Python and its TensorFlow package %  \textit{GradientDescentOptimizer} 
for implementing the proposed approach. 
The results are shown in Table~\ref{natural_situation}. The table compares the results of using individual loss functions and the ensemble loss.
\begin{table}
\begin{tabularx}{\columnwidth}{X XXXX} 
\toprule
 \bf Dataset & \bf Cross-entropy & \bf Hinge &\bf Square &\bf Ensemble\\ 
 \midrule
 20-newsgroups & 0.80 & 0.69 &0.82 &\bf 0.85\\  
 Movie-review &  0.83 & 0.81 & \bf 0.85 & 0.83 \\
  Email-Classification (TREC)& 0.88 & 0.78   & 0.96& \bf 0.97\\
   Reuters& 0.79 & 0.79   & \bf 0.81&\bf 0.81  \\ 
 \bottomrule
\end{tabularx}
\caption{\label{natural_situation} Accuracy}
\end{table}
\begin{table}[ht]
\begin{tabularx}{\columnwidth}{X XXXX} 
\toprule
 \bf Dataset & \bf Cross-entropy & \bf Hinge &\bf Square&\bf Ensemble\\ 
 \midrule
 20-newsgroups & 0.79 & 0.67 &0.69&\bf 0.83\\  
 Movie-reviews & 0.75 & 0.74   & 0.73 & \bf 0.78\\
  Email-Classification (TREC)&0.86  & 0.57   &0.82 &\bf 0.96\\
   Reuters&  \bf 0.76 & 0.69   &0.71& 0.73\\
 \bottomrule
\end{tabularx}
\caption{\label{10 percent outlier} Accuracy in data with 10$\%$  label noise }
\end{table}

We have also compared the robustness of the proposed loss function with the use of individual loss functions. In particular, we add label noise by randomly modifying the target label in the training samples, and keep the evaluation set intact. We conducted experiments with 10\% and 30\% of noise, where e.g. 30\% of noise means randomly changing 30\% of the labels in the training data. Tables~\ref{10 percent outlier} and~\ref{30 percent outlier} show the results, with the best results shown in boldface. We can observe that, in virtually all of the experiments, the ensemble loss is at least as good as the individual losses, and in only two cases the loss is (slightly) worse. And, in general, the ensemble loss performed comparatively better as we increased the label noise.

\section{Conclusion}\label{sec:conclusions}
This paper proposed a new loss function based on ensemble methods. This work focused on text classification tasks and can be considered as an initial attempt to explore the use of ensemble loss functions. %We utilized a softmax neural network and trained the optimal weight of each loss function through the gradient descent mechanism of the neural network. 
The proposed loss function shows an improvement when compared with the use of well-known individual loss functions. Furthermore, the approach is more robust against the presence of label noise. Moreover, according to our experiments, the gradient descent method quickly converged. 

\begin{table}[!htb]
\begin{tabularx}{\columnwidth}{X XXXX} 
\toprule
 \bf Dataset & \bf Cross-entropy & \bf Hinge &\bf Square&\bf Ensemble\\ 
 \midrule
 20-newsgroups & 0.57 & 0.64 & 0.55&\bf 0.82\\  
  movie-review& 0.55 & 0.54   & 0.55&\bf 0.6\\
   Email-Classification (TREC)& 0.80 & 0.46   &0.81&\bf 0.93 \\
   Reuters & 0.64 & 0.54   &0.53& \bf 0.68 \\
 \bottomrule
\end{tabularx}
\caption{\label{30 percent outlier} Accuracy in data with 30$\%$  label noise }
\end{table}

We have used a very simple neural architecture in this work but in principle this method could be used for systems that use any neural networks. In future work we will explore the integration of more complex neural networks such as those using convolutions and recurrent networks. We also plan to study the application of this method to other tasks such as sequence labeling (e.g. for NER and PoS tagging). Another possible extension could focus on handling sparseness by adding a regularization term.

\bibliography{references.bib}
\bibliographystyle{acl_natbib}

\end{document}